\documentclass[runningheads]{llncs}   

\usepackage{eccv}
\usepackage{eccvabbrv}

\usepackage{graphicx}
\usepackage{booktabs}
\usepackage{float}
\usepackage[accsupp]{axessibility}
\usepackage{hyperref}
\usepackage{colortbl}
\usepackage{xcolor}
\usepackage{orcidlink}

\makeatletter
\renewcommand{\@fnsymbol}[1]{%
  \ifcase#1
  \or *%
  \or \dagger%
  \or \ddagger%
  \or \mathsection%
  \or \mathparagraph%
  \else \@ctrerr
  \fi
}
\makeatother

\usepackage{amssymb}  

\makeatletter
\renewcommand{\@fnsymbol}[1]{%
  \ifcase#1\or *\or \textdagger\or \textdaggerdbl
  \or \textsection\or \textparagraph
  \else\@ctrerr\fi}
\makeatother

\begin{document}

\title{VTI-CoT: Visual-Textual Interleaved Chain of Thought for Video Reasoning}

\titlerunning{VTI-CoT}

\author{
Shufan Zhang\inst{1}\thanks{Equal contribution.} \and
Ziyue Lin\inst{2}\protect\footnotemark[1] \and
Bairun Wang\inst{3}\thanks{Corresponding author.} \and
Lei Jin\inst{1} \and
Xuanding Ding\inst{4} \and
Xinzhu Ma\inst{5} \and
Kunlin Yang\inst{3}
}

\authorrunning{S.~Zhang, Z.~Lin et al.}

\institute{
Beijing University of Posts and Telecommunications \\
\email{\{shufanzhang,jinlei\}@bupt.edu.cn}
\and
University of Hong Kong \\
\email{ziyue\_lin@connect.hku.hk}
\and
Beijing Shanwei Zhixing Technology Co., Ltd. \\
\email{\{bairunwang,kunlinyang\}@swellai.net}
\and
Tsinghua University \\
\email{dingxuan@tsinghua.edu.cn}
\and
Beihang University \\
\email{xinzhuma@buaa.edu.cn}
}

\maketitle


\begin{abstract}
Video reasoning aims to understand complex temporal events and causal relationships within videos. Recently, Chain-of-Thought (CoT) has been introduced to this field to enhance reasoning accuracy. However, existing CoT-based video reasoning methods primarily rely on text-only information for logical deduction, overlooking critical visual information during the inference process. Inspired by the human cognitive mechanism of ``reviewing visual segments during inference'', \textbf{we propose VTI-CoT, a Visual-Textual Interleaved CoT Framework}. VTI-CoT integrates textual reasoning steps with its corresponding visual frames. Given the scarcity of visual-textual interleaved CoT in existing datasets, we develop an automated annotation pipeline to construct high-quality multimodal CoT data. Further, reasoning over long-form videos entails increasingly lengthy CoT tokens, which severely hinders training convergence and efficiency. To address this, we employ Contextual Optical Compression(OCR) techniques to compress the CoT supervision signals into a single canvas. Experimental results demonstrate that VTI-CoT achieves state-of-the-art (SOTA) performance among models of the same parameter scale while significantly improving training efficiency.
\keywords{Video Reasoning \and Chain of Thought}
\end{abstract}

\section{Introduction}
\label{sec:intro}
Video reasoning stands as a fundamental task in the field of multi-modal artificial intelligence. It demands that models not only understand image frames but also localize key information and perform logical deduction based on specific instructions. Traditional end-to-end approaches\cite{VideoLLaMA2,MoReVQA,VidCtx} typically map frame features extracted by visual encoders directly to the answer space. Recently, Chain-of-Thought (CoT) \cite{Chain-of-Thought,zhou2023leasttomostpromptingenablescomplex,xiang2025} has been introduced to this domain. By decomposing complex reasoning tasks into interpretable intermediate steps, CoT significantly enhances both the interpretability and accuracy of video reasoning.

When human beings make inferences from a video, the reasoning process is often interleaved with the review of specific video segments. However, existing CoT-based video reasoning methods \cite{VideoLLaMA2,Video-R1,VideoCoT} treat visual inputs and reasoning chains as disconnected streams. While models process the visual information, their reasoning remains purely text-centric, lacking explicit integration between textual rationales and specific visual evidence\cite{Video-R2,V-STaR}. This prevents the model from periodically reviewing relevant visual information to validate the correctness of its logical reasoning process\cite{TimeChat}. Some works \cite{Chain-of-Frames} attempt to link the reasoning process with corresponding frames, but only by adding simple frame indices to the text. Consequently, current Video Large Language Models (Video-LLMs) forget key information when conduct logical reasoning, which is a waste of the models' powerful reasoning abilities. 



To address this, we propose a novel Visual-Textual Interleaved video reasoning framework, named as VTI-CoT. Our approach localizes the most relevant video segments corresponding to the problem conditions and reasoning steps, and constructs a CoT that integrates video descriptions, visual information (images), and the model's reasoning process. Such a CoT construction and reasoning strategy effectively improves the model's ability to revisit key information during inference, further enhancing the accuracy of the final answers. Current widely-used datasets fail to satisfy the requirements for explicit visual-textual information in video reasoning, thereby we develop an automated pipeline to curate a visual-textual interleaved CoT dataset. Specifically, we first partition videos into coherent segments based on semantic similarity. Subsequently, we leverage advanced Large Language Models (LLMs) to synthesize corresponding textual descriptions for each visual segment, and then generate multi-step reasoning chains.


After generating rich multi-modal chain-of-thought information, it is necessary to supervise a large amount of tokens during model training. In case of reasoning chains that address complex questions in long videos, it will lead to an explosion number of tokens. This results in significant memory consumption and an increased difficulty in model training. Inspired by DeepSeek-OCR\cite{DeepSeek-OCR}, we leverage document-level Contexts Optical Compression(OCR) representations to densely encode textual and visual content. Specifically, we render interleaved text-image CoT data into a canvas and extract features from this canvas through a general vision encoder\cite{qwen2025qwen25}. These features are used as supervision signals during model training, so that we can train CoT with a fixed number of tokens. This training approach significantly accelerates convergence.

We evaluate our method across multiple datasets and achieve superior results. Notably, our approach demonstrates exceptional performance comparing to current state-of-the-art methods.

In a conclusion, we introduce VTI-CoT, a novel Chain-of-Thought framework with both text and visual information. It generates structured reasoning chains, achieving mutual complementarity between textual rationale and visual information review. We apply an OCR-based rendering method during training, which renders interleaved text-image CoT data into a canvas. It can effectively compress reasoning tokens, enhance information density and thereby improve training efficiency. Our method shows state-of-the-art performance and training efficiency across multiple video understanding and reasoning benchmarks.

\section{VTI-CoT: Visual-Textual Interleaved Chain of Thought}

We propose a Visual-Textual Interleaved Chain-of-Thought (VTI-CoT) framework for video reasoning. Unlike conventional CoT supervision that relies purely on textual reasoning, our method additionally identifies the key video segments corresponding to each reasoning step, using them as supplementary information or visual evidence to enhance the completeness of the reasoning process.
Current datasets cannot fully support our training requirements, and manual annotation would be prohibitively expensive. 

To address this, we first design an automated pipeline to construct structured CoT supervision from raw videos using large language models. 
Furthermore, instead of supervising long reasoning chains token by token, we convert the structured multi-modal CoT into a compact visual representation and train the model to predict this representation. This strategy improves training efficiency while maintaining the integration of textual reasoning steps and their associated video evidence.
Our construction pipeline contains four stages: \textbf{(1) Temporal segmentation, (2) Interval-level description, (3) Answer-conditioned reasoning generation, and (4) OCR rendering.} See complete pipeline in Fig.~\ref{fig:pipeline}

\subsection{Visual-Textual Interleaved CoT Construction}
Large language models show strong ability in multi-step reasoning\cite{GPT-4,PaLM,TreeofThoughts} . We exploit this ability to construct CoT supervision for long videos without manual annotation. Instead of letting the model reason freely over the entire video, we force each reasoning step to be tied to a specific temporal region and its visual evidence. 
\paragraph{\textbf{1. Temporal Segmentation.}}
Our framework begins by identifying key temporal boundaries to capture the event structure of the long-form video. Given a raw video, we uniformly sample 256 frames to balance coverage and efficiency\cite{VideoMAE,VideoSWin}. 
For each sampled frame $f_i$, we first extract a visual feature $\phi(f_i)$ using CLIP\cite{clip} encoder (ViT-B/32), and then compute cosine similarity between adjacent frames:
\begin{equation}
s_i = 
\frac{\phi(f_i) \cdot \phi(f_{i+1})}
{\|\phi(f_i)\| \, \|\phi(f_{i+1})\|}, 
\quad i = 1, \dots, 255.
\end{equation}
\noindent where low similarity indicates a visual transition. We rank all similarity values and select the $K$ largest drops (default $K = 20$) as temporal boundaries. This approach partitions the video into a sequence of contiguous intervals, where each interval roughly corresponds to a short, semantically coherent segment\cite{Revisiting}. This segmentation provides the temporal information for reasoning.
\begin{figure}[t]
    \centering
    \includegraphics[width=0.93\textwidth]{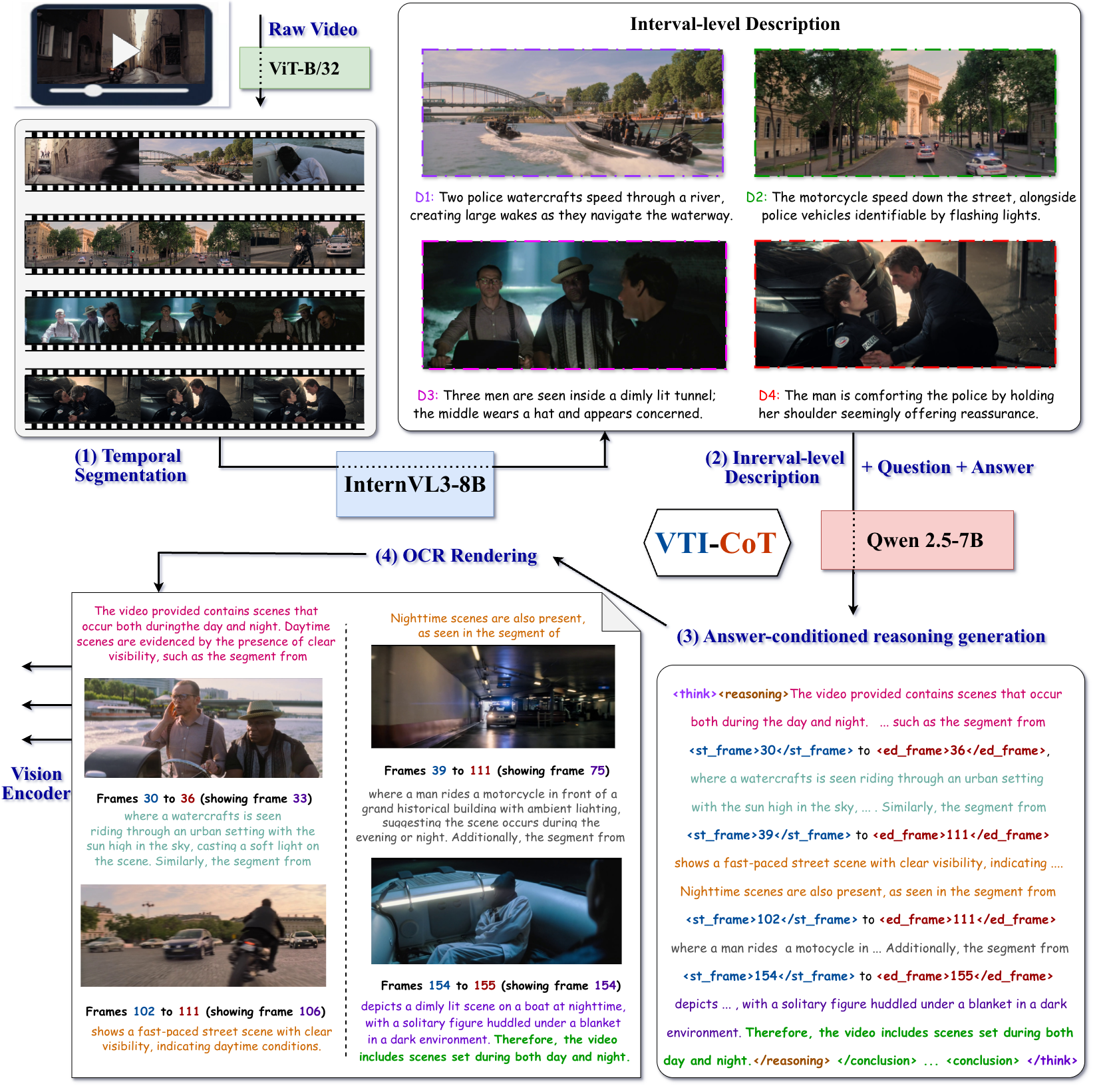}
    \caption{\textit{VTI-CoT data generation pipeline.} Our construction pipeline contains four stages: (1) Temporal segmentation: Segments video frames based on CLIP. (2) Interval-level description: Generate detailed segment descriptions. (3) Answer-conditioned reasoning generation: Generate visual-textual interleaved CoT. (4) OCR rendering: Write CoT into a canvas, then feed the canvas into a vision encoder to generate tokens.}
    \label{fig:pipeline}
\end{figure}

\paragraph{\textbf{2. Interval-level visual description.}}
For each interval $[i_k, i_{k+1})$, we use InternVL3-8B\cite{chen2024internvl} to generate a concise description $d_k$. As illustrated in Fig.~\textbf{~\ref{fig:pipeline}}, each segmented interval is summarized into a structured textual description\cite{InternVid}. Each description follows the format $(i_k, i_{k+1}) : d_k$, where $d_k$ summarizes the main entities, actions, and visual changes inside the interval. 
The collection $\{(i_k, i_{k+1}) : d_k\}_{k=1}^{K}$ forms a structured textual abstraction of the full video\cite{STEP}. These interval descriptions are the only visual abstraction given to the reasoning model at this stage. It is worth noting that the model never observes the full video holistically, which prevents ungrounded global inference\cite{Liang_2023}.
\paragraph{\textbf{3. Answer-conditioned reasoning generation.}}
Given the interval descriptions $\{(i_k, i_{k+1}) : d_k\}_{k=1}^{K}$ together with the question and ground-truth answer, we prompt Qwen2.5-7B\cite{qwen2025qwen25} to generate a step-wise reasoning chain. Each reasoning step is required to explicitly cite its supporting interval in the form $r_t \;|\; [i_{ts}, i_{te}]$, for $t = 1, \dots, T$. This constraint enforces temporal grounding at every reasoning step and prevents the model from generating free-form global explanations without evidence reference\cite{LeAdQA} .

Specifically, the prompt enforces a structured XML output comprising three distinct components:
(i) a step-by-step reasoning trajectory, where each reasoning step is explicitly grounded to a temporal interval;
(ii) a global conclusion grounded in specific temporal segments via $<st\_frame>$ and $<ed\_frame>$ tags; and
(iii) a concise final answer.
The resulting output defines a structured CoT instance with three hierarchically organized components:
(1) a temporally grounded reasoning trajectory $\{r_t\}_{t=1}^{T}$, where each reasoning step explicitly cites its supporting interval;
(2) a global reasoning summary $\mathcal{G}$ that aggregates cross-interval evidence and provides a structured conclusion grounded by explicit temporal tags (e.g., $<st\_frame>$ and $<ed\_frame>$); and
(3) a concise final answer derived from the accumulated reasoning process.The detailed generation prompt is provided in Appendix~\ref{appendix:dataset}.

\subsection{Interleaved CoT Rendering for Compact Visual Supervision}

The structured reasoning instance produced in the previous steps explicitly specifies which temporal interval supports each reasoning step. We now convert this temporally grounded CoT into a compact visual supervision signal by extracting visual evidence and rendering an interleaved image-text reasoning chain.

\paragraph{\textbf{4. OCR rendering.}}
For each referenced interval $[i_s, i_e]$ in the reasoning trajectory, we deterministically retrieve supporting frames from the original video. As illustrated in Fig.\textbf{~\ref{fig:pipeline}}, this process pairs each reasoning step with its visual evidence in a structured manner. If $i_s = i_e$, we retain the corresponding single frame. Otherwise, we extract the midpoint of the interval. 
Let $V_t$ denote the frame set associated with reasoning step $r_t$.
This yields a structured instance
$\mathcal{C} = \{ (r_t, V_t)_{t=1}^{T}, \mathcal{G} \}$,
where each reasoning step is explicitly paired with its visual evidence,
and $\mathcal{G}$ denotes the global reasoning summary.
Directly supervising long reasoning chains at the token level leads to linearly growing sequence length, increased computational cost, and training instability. To obtain a compact and unified supervision signal, we render the structured instance into a single RGB canvas:
\begin{equation}
I_{\mathrm{CoT}} = Render(\{(r_t, V_t)\}_{t=1}^{T}; \psi),
\end{equation}
where $\psi$ specifies rendering hyperparameters such as font size, padding, spacing, and layout bounds~\cite{DeepSeek-OCR,Render-of-Thought,DeepSeek-OCR2}.

The renderer arranges textual rationales and their associated frames sequentially along the horizontal axis. While the image width remains constant, the canvas height scales dynamically to accommodate the total number of reasoning steps. 
Specifically, each textual rationale $r_t$ is placed adjacent to its associated visual set $V_t$. This interleaved layout of text and image allows the model to revisit key visual information while processing the reasoning trajectory. We provide an example of the rendered CoT canvas in Appendix~\ref{appendix:dataset}.

The resulting CoT canvas is processed by the vision encoder of Qwen2.5-VL-7B-Instruct~\cite{qwen2025qwen25}, producing a compressed optical signal that serves as the supervisory input for training the reasoning model. This compression reduces the computational cost required to process the reasoning trajectory~\cite{InternVid}. Empirically, this design accelerates training efficiency. The rendering process itself is fully deterministic and parameter-free.

\begin{table}[b]
\centering
\caption{Video Reasoning Datasets Overview}
\label{data1}
\resizebox{\textwidth}{!}{
\begin{tabular}{c|c|c|c|c} 
\toprule
\textbf{Data Source} & \textbf{Count} & \textbf{Avg. Tokens} & \textbf{Avg. Images} & \textbf{Introduction} \\
\midrule
Video-R1 & 86,217 & 1,406 & 3.91 & Short video clips, multi-step reasoning \\
MovieChat & 4,136 & 3,439 & 10.72 & Long narrative, temporal/causal deps \\
\bottomrule
\end{tabular}
}
\end{table}

\begin{figure}[t]
    \centering
    \includegraphics[width=0.85\textwidth]{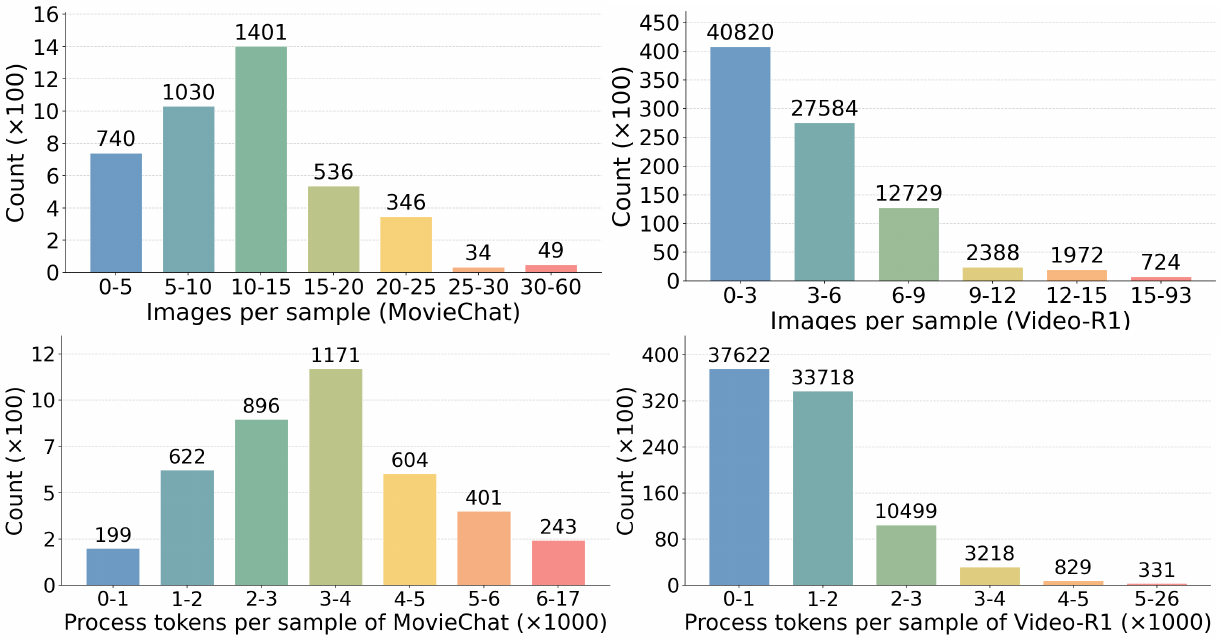}
    \caption{
    \textit{Dataset statistics of Video-R1 and MovieChat}, including number of tokens and images per video sample.
    }
    \label{fig:data_statistics}
\end{figure}
\subsection{Dataset Synthesis}
\label{dataset synthesis}
Most existing dataset don't fully satisfy the requirements for visual-textual interleaved CoT reasoning. Hence we establish and apply the VTI-CoT construction pipeline to two popular video-reasoning datasets (See Fig.~\ref{fig:data_statistics}):
\begin{itemize}
    \item 1.\textbf{VTI-}Video-R1-CoT-165K: Built upon Video-R1-CoT-165K~\cite{Video-R1}, this dataset is constructed using our proposed automated pipeline. It consists of 86,217 videos annotated with temporally grounded CoT supervision. Each sample contains multi-step reasoning explicitly aligned with localized video segments, providing fine-grained supervision for long-video reasoning.
    
    \item 2.\textbf{VTI-}MovieChat: Built upon MovieChat\cite{MovieChat}, this dataset is constructed using our proposed automated pipeline. MovieChat\cite{MovieChat} contains 4,136 long-form movie videos without explicit CoT annotations, primarily provides diverse narrative-style videos with complex temporal structures.
    
\end{itemize}



Fig.~\ref{fig:data_statistics} and Tab.~\ref{data1} summarize the statistics of VTI-Video-R1-CoT-165K and VTI-MovieChat in terms of scale and structural properties. For each video sample, \textbf{Avg. Tokens} denotes the average length of the reasoning chain after tokenizing both the textual reasoning steps and their associated frames, reflecting the overall tokenized information processed per video sample. \textbf{Avg. Images} indicates the average number of frames included in each rendered canvas, capturing the visual volume for each video sample in our VTI-CoT. As illustrated in Fig.~\ref{fig:data_statistics}, the CoT sequences in both datasets not only exhibit a high token number but also significant variance in their distribution, which poses a substantial challenge for stable model training.

MovieChat is mainly composed of long videos, which leads to more reasoning tokens and more keyframes per reasoning step. In contrast, Video-R1 consists mostly of shorter videos with more concise reasoning tokens and fewer keyframes per step. The distribution of video types further indicates that Video-R1 emphasizes temporal and causal reasoning, whereas MovieChat focuses on narrative-level long-video understanding.
\begin{figure}[htbp] 
    \centering 
    \includegraphics[width=0.99\textwidth]{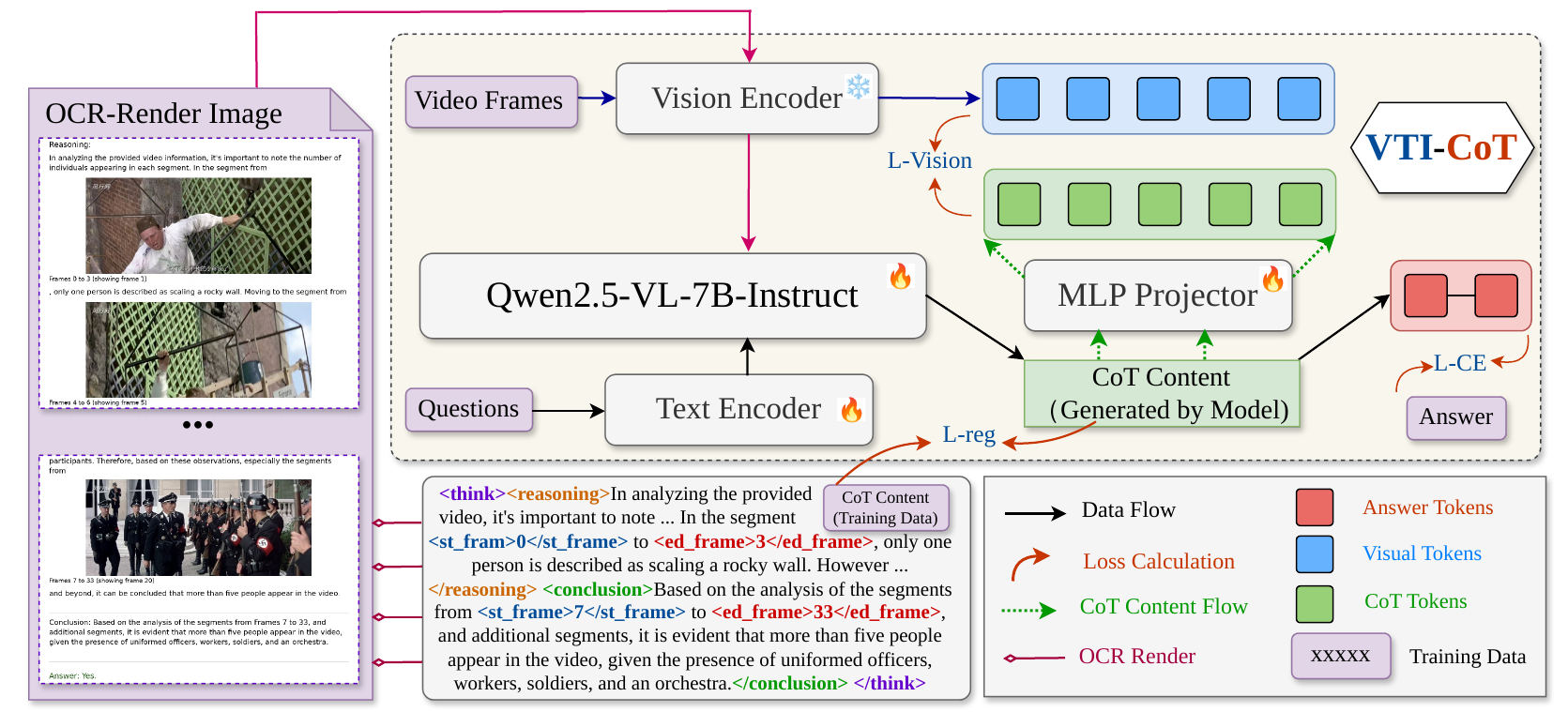} 
    \caption{\textit{Proposed VTI-CoT training framework.} VTI-CoT integrates visual and textual information into an encoded feature by utilizing OCR method. In training stage, we first render image-text interleaved CoT content into a canvas, then encode this image through a general vision encoder, finally integrating this feature with CoT content generated by LLM.} 
    \label{fig:main} 
\end{figure}
\subsection{Training with Token-Level Interleaved CoT Supervision}
We adopt Qwen2.5-VL-7B-Instruct\cite{qwen2025qwen25} as our base model and follow the training process on Video-R1's\cite{Video-R1} official implementation. Hyperparameters and detailed configurations are provided in Appendix~\ref{appendix:train}. We will explain how we train this VTI-CoT framework in below and we also show the process in Fig.~\ref{fig:main}.

\paragraph{\textbf{1. Vision-aware CoT representation.}}
Unlike conventional multimodal LLMs that inject visual patch tokens into the input stream, we incorporate visual features into the language model’s hidden states~\cite{Chain-of-Visual-Thought,SEA}. This design enables the model to internalize visual evidence during training while preserving standard autoregressive decoding at inference.

For each interleaved visual-textual CoT instance constructed in \ref{dataset synthesis}, we structure the model’s output sequence to separately account for textual reasoning, visual tokens, and conclusion:
\begin{equation}
\mathcal{C} = \big[ \mathbf{T}_{\mathrm{re}}, \; \mathbf{a}_1, \dots, \mathbf{a}_{N_v}, \; \mathbf{T}_{\mathrm{co}} \big],
\end{equation}
where $\mathbf{T}_{\mathrm{re}}$ and $\mathbf{T}_{\mathrm{co}}$ denote the textual reasoning and conclusion segments, and $\{\mathbf{a}_i\}_{i=1}^{N_v}$ is a fixed-length block representing visual tokens. The number of visual tokens $N_v$ matches the number of patch-level embeddings extracted by a frozen vision encoder.

For each rendered CoT canvas, we first resize it to a fixed resolution of $N_v \times 28 \times 28$ and feed it into the vision encoder. The encoder produces patch-level embeddings 
$\mathbf{V} = \{\mathbf{v}_i\}_{i=1}^{N_v}$, where each $\mathbf{v}_i$ corresponds to 
During training, we collect the hidden states at the visual token positions, forming 
$\mathbf{H} \in \mathbb{R}^{N_v \times d}$, 
where $d$ is the hidden dimension of the language model. A lightweight projection head $\phi(\cdot)$ maps $\mathbf{H}$ into the visual feature space, yielding predicted embeddings $\hat{\mathbf{V}} = \phi(\mathbf{H})$. We compute the mean squared error between $\hat{\mathbf{V}}$ and $\mathbf{V}$:
\begin{equation}
\mathcal{L}_{\mathrm{vision}} =
\frac{1}{N_v}
\sum_{i=1}^{N_v}
\left\|
\hat{\mathbf{v}}_i - \mathbf{v}_i
\right\|_2^2 .
\end{equation}

\paragraph{\textbf{2. Overall Training Objective.}}
In addition to visual supervision, we apply standard autoregressive language modeling over the full CoT sequence. Let $Y = [y_1, y_2, \dots, y_T]$ denote the target token sequence of the CoT instance (including reasoning steps and the final answer). The cross-entropy loss is defined as:
\begin{equation}
\mathcal{L}_{\mathrm{CE}} =
- \sum_{t=1}^{T} \log P_{\theta}(y_t \mid y_{<t}),
\end{equation}

\noindent where $y_t$ is the $t$-th token in the output CoT sequence and $y_{<t}$ represents all preceding tokens.
To maintain the structural integrity of the alignment region, we include a lightweight regularization term that enforces the presence of the boundary markers $<cot\_start>$ and $<cot\_end>$ in the output sequence $\mathcal{C}$. Formally:
\begin{equation}
\mathcal{L}_{\mathrm{reg}} =
\mathbf{1}_{\{<cot-start> \notin \mathcal{C} \;or\; <cot-end> \notin \mathcal{C}\}},
\end{equation}
where $\mathcal{C}$ denotes the full output sequence. This regularization ensures that each CoT instance has clearly defined start and end boundaries, which stabilizes training.
The total training objective is therefore:
\begin{equation}
\mathcal{L}_{\mathrm{total}}
=
\lambda_{\mathrm{LM}} \mathcal{L}_{\mathrm{CE}}
+
\lambda_{\mathrm{vision}} \mathcal{L}_{\mathrm{vision}}
+
\lambda_{\mathrm{reg}} \mathcal{L}_{\mathrm{reg}},
\end{equation}

\noindent where $\lambda_{\mathrm{LM}} = 1, \lambda_{\mathrm{vision}} = 0.5, \lambda_{\mathrm{reg}} = 1$ in our experiment setting.


\section{Experiments}

\subsection{Setup}

\paragraph{\textbf{1. Training and Evaluation.}} 
All experiments are based on Qwen2.5-VL-7B-Instruct~\cite{qwen2025qwen25}, which serves as the multimodal backbone for our framework.We perform supervised fine-tuning (SFT) on the video reasoning datasets constructed with the VTI-CoT pipeline (Section~\ref{dataset synthesis}).

For efficiency, the maximum number of video frames is limited to 16 during training, and each frame is processed at a resolution of $128 \times 28 \times 28$ pixels. We optimize the model for one epoch with a learning rate of $3 \times 10^{-6}$ and a global batch size of 1. Unless otherwise specified, all ablation variants share the same initialization, training schedule, and optimization settings. During inference, we may extend the maximum frames to 32 with a maximum resolution of $256 \times 28 \times 28$.


\paragraph{Benchmarks.}
We evaluate our method on six widely used video benchmarks that collectively assess both general video understanding and long-video reasoning:

\begin{itemize}
    \item Video general benchmarks:  
     I. MVBench~\cite{mvbench}: Evaluates temporal ordering, action understanding, and multi-modal reasoning on relatively short video clips.
    II. TempCompass~\cite{TempCompass}: Focuses on temporal reasoning and event sequencing over moderately long inputs.   
    III. Video‑MME~\cite{videomme}: A comprehensive multimodal evaluation with hundreds of videos and thousands of QA pairs, designed to measure broad video comprehension skills including subtitle and audio usage.
    IV. MMVU~\cite{MMVU}: Evaluates general video understanding with multiple-choice questions for consistency and stability. 
    \item Long-video benchmarks:
    V. LVBench~\cite{LVbench}: Tests understanding over long-duration videos with sparse visual evidence, requiring multi-step temporal reasoning. 
    VI. LongVideoBench~\cite{LongvideoBench}: Specifically designed for interleaved long-context video-language understanding, with questions dependent on retrieving and reasoning over detailed information across extended video segments. 
\end{itemize}

\subsection{Main Results}

\paragraph{Comparison with state-of-the-art models.} 

We evaluate our method against a diverse set of video reasoning models across four general and two long-video understanding benchmarks. As summarized in Tab.~\ref{tab:video_benchmark_redesign}, our method consistently achieves state-of-the-art performance.
We first evaluate the effectiveness of our reasoning framework under a controlled setting using the same training data. Compared with the video-only baseline Video-R1-wo-image~\cite{Video-R1}, which is trained on the identical dataset as Ours(wo-mv), our model achieves consistent improvements across all benchmarks.
When additional training data are incorporated, the full model further improves performance. As shown in the final row of Tab.~\ref{tab:video_benchmark_redesign}, our method consistently surpasses the multimodal Video-R1 across all general video benchmarks.

These results demonstrate the effectiveness of our method: by explicitly grounding intermediate rationales in visual evidence, the framework enables multi-step inference that is more structured and reliable.

\begin{table*}[t]
\centering
\caption{Performance comparison across video general and long-video benchmarks.}
\label{tab:video_benchmark_redesign}
\resizebox{\textwidth}{!}{
\begin{tabular}{l c c cccc cc}
\toprule
\textbf{Model} & \textbf{Size} & \textbf{\#Frames} 
& \multicolumn{4}{c}{\textbf{Video General Benchmarks}} 
& \textbf{Long} & \textbf{Extremely Long} \\
\cmidrule(lr){4-7} \cmidrule(lr){8-9}
& & 
& \textbf{MVBench} & \textbf{TempCompass} & \textbf{Video-MME} & \textbf{MMVU(mc)} 
& \textbf{LongVideoBench} & \textbf{LVBench} \\
\midrule

\rowcolor{gray!15} \multicolumn{9}{c}{\textbf{Proprietary MLLMs}} \\
GPT-4o & -- & -- & 73.8 & 71.9 & 75.4 & -- & 66.7 & -- \\
GPT-5 & -- & -- & 83.3 & 86.7 & 82.6 & -- & -- & -- \\
Gemini-1.5-Pro & -- & -- & 67.1 & 75.0 & 71.2 & -- & 64.0 & -- \\
Gemini-2.5-Pro & -- & -- & 84.3 & 84.3 & 78.4 & -- & -- & -- \\
\midrule
\rowcolor{gray!15} \multicolumn{9}{c}{\textbf{Open-Source MLLMs}} \\
LLaVA-OneVision & 7B & 64 & 56.7 & 64.2 & 58.2 & 49.2 & 42.6$^{*}$ & 38.4$^{*}$ \\
VideoLLaMA2 & 8$\times$7B & 16 & 53.9 & -- & 47.9 & 44.8 & 36.0 & -- \\
LongVA & 7B & 128 & 56.9 & -- & 52.6 & -- & 47.8 & -- \\
Video-XL & 7B & 128 & 55.3 & -- & 55.5 & -- & 50.7 & -- \\
Kangaroo & 8B & 64 & 61.1 & 62.5 & 56.0 & -- & 54.8 & 39.4 \\
VideoChat-R1 & 7B & 16 & 63.1 & 72.9 & 52.4 & 64.2 & 53.3 & 34.1 \\
Qwen2.5-VL-7B & 7B & --  & 59.4 & 72.5 & 56.5 & 59.2 & 49.1 & 32.6 \\
Qwen2.5-VL-7B-SFT & 7B & 32 & 60.5 & 69.9 & 55.4 & 63.5 & 51.2$^{*}$ & 35.7$^{*}$ \\
\midrule
\rowcolor{gray!15} \multicolumn{9}{c}{\textbf{SOTA Models}} \\
Video-R1 & 7B & 32 & 63.9 & 73.2 & 59.3 & 63.8 & 54.6 & 38.7 \\
Video-R1-wo-image & 7B & 16 & 60.9 & 69.8 & 53.8 & 60.6 & 51.5$^{*}$ & 34.7$^{*}$ \\
Video-R1-wo-image & 7B & 32 & 63.2$^{*}$ & 71.8$^{*}$ & 57.2$^{*}$ & 62.7$^{*}$ & 53.4$^{*}$ & 35.6$^{*}$ \\
Ours(wo-mv) & 7B & 32 & 65.2 & 73.8 & 59.1 & 63.8 & 54.8 & 40.4 \\

\midrule
\textbf{Ours} & 7B & 32 & \textbf{65.9} & \textbf{74.5} & \textbf{59.6} & \textbf{65.3} & \textbf{55.0} & \textbf{40.5} \\
\bottomrule
\multicolumn{9}{l}{\small{$^{*}$ Evaluated by ourselves.}}
\end{tabular}
}
\end{table*}

\subsection{Ablation Studies}




To systematically analyze the contribution of each component in our framework,
we conduct a series of ablation experiments.
Unless otherwise specified, all variants use Qwen2.5-VL-7B-Instruct as the base model and share identical data, optimization protocols, and training schedules.

\paragraph{\textbf{1.Influence of Adding Visual Information in Video CoT Reasoning.}}
This experiment aims to validate the influence of visual information in Chain-of-Thought video (CoT) reasoning.
We compare three variants of our framework: 
(1) \textbf{Interleaved CoT}, which contains both textual reasoning steps and their corresponding visual evidence; 
(2) \textbf{Text-only CoT}, which removes all visual information and retains only the textual reasoning steps; 
(3) \textbf{Image-only CoT}, which removes textual reasoning, leaving only the visual inputs. 

The results are presented in Tab.~\ref{tab:exp1_multimodal}, where we observe that removing either modality(text or visual) consistently leads to performance degradation across all benchmarks. 
Specifically, Text-only CoT lacks explicit visual cues, which limits the model's ability to review critical frames and assemble evidence across multiple reasoning steps. 
Conversely, Image-only CoT lacks logical structure and reasoning signals, resulting in an inability to perform complex multi-step reasoning despite having access to visual information. 
These results demonstrate that textual and visual modalities are complementary and essential for video multi-step reasoning. Interleaving reasoning steps with their corresponding visual evidence enables the model to learn coherent trajectories that capture temporal, causal, and semantic dependencies.

\begin{table}[t]
\centering
\caption{\textit{Ablation study on textual and visual components in interleaved CoT.} 
We compare text-only, image-only, and fully interleaved CoT variants to assess the contribution of each modality to multi-step video reasoning performance.}
\label{tab:exp1_multimodal}
\resizebox{\linewidth}{!}{
\begin{tabular}{l|c|c|c|c|c|c}
\toprule
\textbf{CoT Variant} 
& MVBench & TempCompass & VideoMME & MMVU & LongVideoBench & LVBench   \\
\midrule
\textbf{Interleaved CoT}
& \textbf{65.9} & \textbf{74.5} & \textbf{59.6} & \textbf{65.3} & \textbf{55.0} & \textbf{40.5}  \\
Text-only CoT
& 65.4 & 74.0 & 59.3 & 64.9 & 54.5 & 40.0   \\
Image-only CoT
& 64.2 & 72.6 & 57.2 & 64.5 & 52.4 & 38.8  \\
\bottomrule
\end{tabular}
}
\end{table}

\begin{table}[t]
\centering
\caption{\textit{Ablation on structured rendering in multimodal CoT.} Tokenized inputs are compared with rendered CoT to highlight the effect of compact supervision on learning efficiency and accuracy.}
\label{tab:exp2_rendered}
\resizebox{\textwidth}{!}{%
\begin{tabular}{l|c|c|c|c|c|c}
\toprule
\textbf{Method} 
& MVBench & TempCompass & VideoMME & MMVU & LongVideoBench & LVBench  \\
\midrule
\textbf{Rendered CoT} 
& \textbf{65.9} & \textbf{74.5} & \textbf{59.6} & \textbf{65.3} & \textbf{55.0} & \textbf{40.5} \\
Tokenized CoT 
& 63.2 & 71.3 & 57.9 & 63.8 & 53.5 & 39.5 \\
\bottomrule
\end{tabular}%
}
\end{table}

\paragraph{\textbf{2. Influence of OCR Rendering in Video CoT Reasoning.}} 
This experiment examines whether performance gains stem not merely from providing multi-modal inputs, but also from rendering the structured reasoning trajectory.

We compare our \textbf{Rendered Interleaved CoT} with a \textbf{Tokenized Multimodal CoT} baseline (without rendering and feature extracting).
In the tokenized setting, textual reasoning steps and their associated images are directly tokenized and concatenated as standard multi-modal inputs. Such formulation provides multi-modal inputs, preserving integration of reasoning text and its corresponding visual cues. However, the resulting token sequences become very long, sparse, and redundant. The increased sequence length and reduced information density make learning more difficult, slowing convergence and lowering final performance.

\begin{figure}[t]
    \centering
    \begin{subfigure}{0.48\textwidth}
        \centering
        \includegraphics[width=\linewidth]{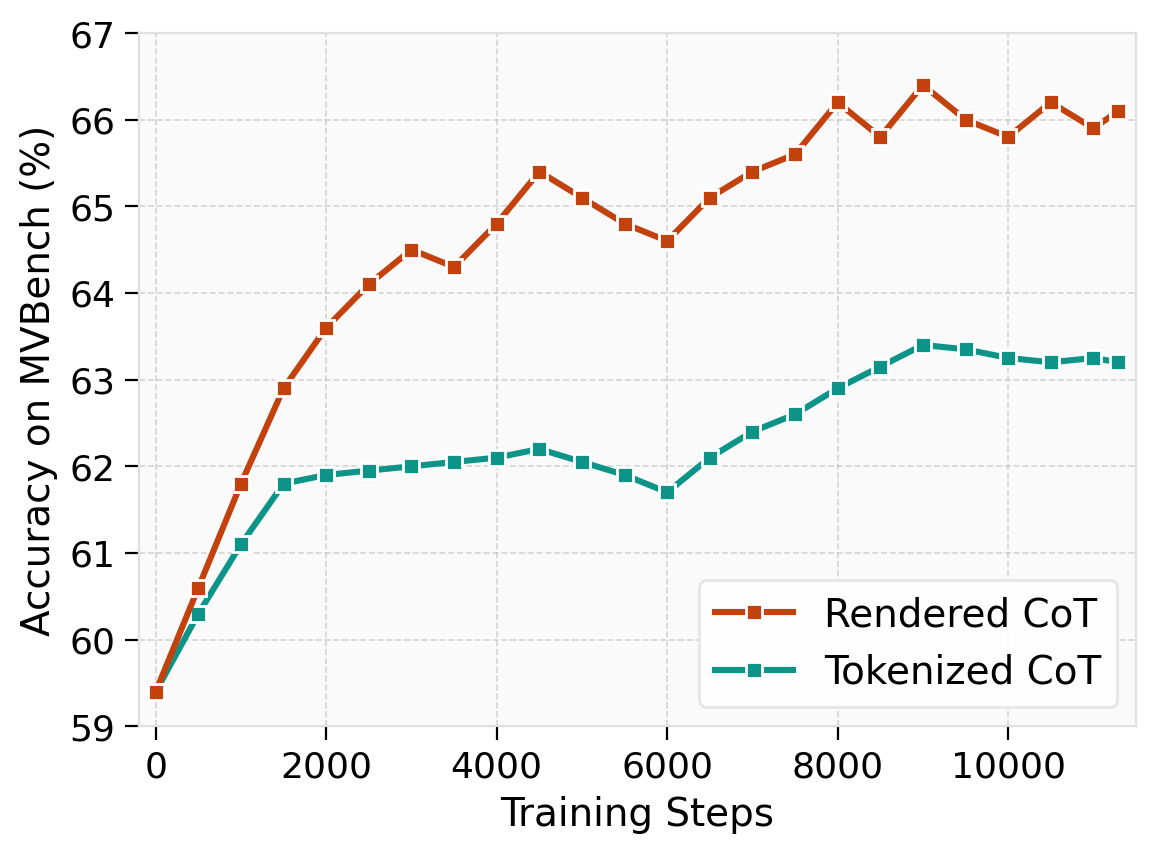}
    \end{subfigure}
    \hfill
    \begin{subfigure}{0.48\textwidth}
        \centering
        \includegraphics[width=\linewidth]{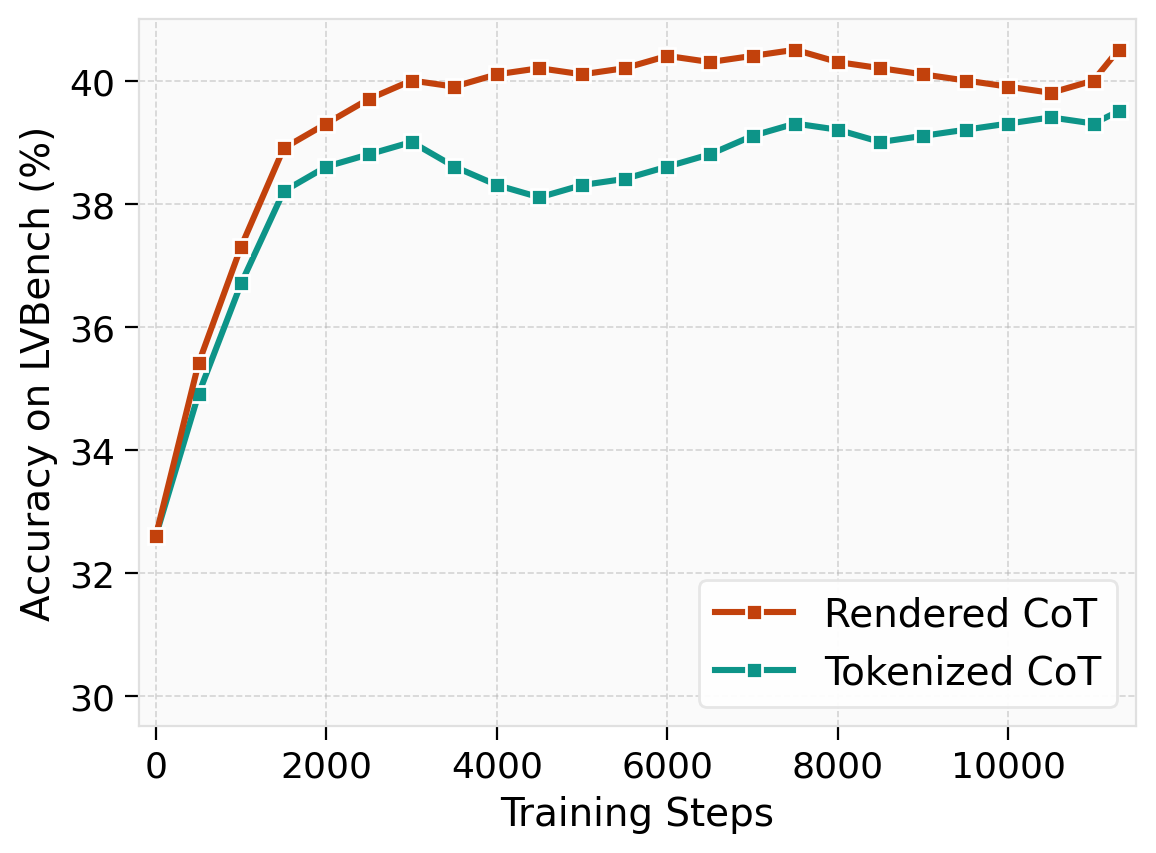}
    \end{subfigure}

    \caption{
    \textit{Training curves of rendered CoT versus tokenized CoT on MVBench and LongVideoBench.} Results show that rendered CoT converges faster tan tokenized CoT in both short and long video benchmarks.}
    
    \label{fig:training}
\end{figure}

In contrast, the rendered CoT converts the structured reasoning trajectory into a compact image-text canvas. 
Each reasoning step is associated with its corresponding visual evidence in a dense, spatially organized format, which improves training efficiency. 
As shown in Tab.~\ref{tab:exp2_rendered}, rendered CoT supervision consistently outperforms tokenized multi-modal CoT across all benchmarks. 
Fig.~\ref{fig:training} further illustrates the training curves of the two approaches. The rendered formulation converges faster and achieves higher final accuracy.
These results indicate that the benefits of our method primarily arise from \emph{structuring and densifying} the reasoning trajectory through rendering, rather than from multi-modal input exposure alone.

\paragraph{\textbf{3. Effect of Reasoning Order Consistency.}}
We examine whether the model truly learns the logical order of the reasoning process.
We compare the correctly ordered \textbf{Ordered CoT} with a \textbf{Shuffled CoT} variant,
where image-text pairs are preserved but randomly permuted along the reasoning trajectory.

As shown in Tab.~\ref{tab:exp3_shuffle}, shuffling the order consistently degrades performance across benchmarks.
This indicates that maintaining the correct temporal and logical alignment between reasoning steps and visual evidence is crucial.
The model benefits not only from multimodal exposure but also from learning a coherent reasoning trajectory grounded in visual evidence.

\begin{table}[t]
\centering
\caption{
\textit{Comparison between correctly ordered interleaved CoT and its shuffled variants.} Results shows that ordered CoT performs better in all benchmarks.}
\label{tab:exp3_shuffle}
\resizebox{\linewidth}{!}{
\begin{tabular}{l|c|c|c|c|c|c}
\toprule
\textbf{CoT Variant} 
& MVBench  & TempCompass & VideoMME & MMVU & LongVideoBench & LVBench   \\
\midrule
\textbf{Ordered CoT}
& \textbf{65.9} & \textbf{74.5} & \textbf{59.6} & \textbf{65.3} & \textbf{55.0} & \textbf{40.5} \\
Shuffled CoT
& 65.2 & 73.9 & 58.3 & 64.8 & 53.8 & 39.9  \\
\bottomrule
\end{tabular}
}
\end{table}

\begin{table}[t]
\centering
\caption{\textit{Ablation study on different vision encoders used for patch-level feature extraction.} We replace the default Qwen vision encoder with CLIP while keeping all other components unchanged.}
\label{tab:exp4}
\resizebox{\linewidth}{!}{
\begin{tabular}{l|c|c|c|c|c|c}
\toprule
\textbf{Vision Encoder} & MVBench & TempCompass & VideoMME & MMVU & LongVideoBench  & LVBench \\
\midrule
\textbf{Qwen Encoder} & \textbf{65.9} & \textbf{74.5} & \textbf{59.6} & \textbf{65.3} & \textbf{55.0} & \textbf{40.5}  \\
CLIP Encoder& 65.5 & 73.8 & 57.5 & 64.3 & 54.7 & 40.2 \\
\bottomrule
\end{tabular}
}
\end{table}

\paragraph{\textbf{4. Qwen Image Encoder Works Better Than CLIP Image Encoder.}}
We further investigate how different image encoder affects the performance of CoT canvas feature extraction. In our default setting, we adopt the original vision encoder from Qwen2.5-VL-7B-Instruct~\cite{qwen2025qwen25}. To study this factor, we replace it with a pretrained CLIP~\cite{clip} image encoder while keeping other components unchanged.

All variants are trained under the same data setting, optimization protocol, and training schedule. The results are shown in Tab.~\ref{tab:exp4}. We observe that using the Qwen image encoder consistently achieves better performance than CLIP across all benchmarks. This suggests that Qwen preserves richer semantic information when encoding the rendered CoT canvas, producing more informative compressed visual signals for reasoning.

These results indicate that a stronger vision encoder helps extract higher-quality CoT representations, enabling the model to better understand the reasoning trajectory and improving inference accuracy. At the same time, the performance gap remains moderate, suggesting that our framework does not rely on a specific encoder architecture. Overall, this demonstrates both the importance of high-quality visual feature extraction and the robustness of our approach.

\section{Related Works}

\subsection{Video Reasoning}
The primary challenge in video reasoning lies in the model's difficulty in integrating cross-modal information during long context inference, where visual cues are frequently neglected by the generated textual rationales. Early works tackles these issues by aggregating segment-level features and modeling temporal dependencies. LongVideoQA~\cite{LongVideoQA} introduces segment-level feature aggregation to support multi-step reasoning over extended videos. TempCompass~\cite{TempCompass} proposes temporal attention mechanisms to emphasize sparse key frames while maintaining global context. Video-MTR~\cite{videomtr} leverages hierarchical video representations and multi-task learning to improve cross-task long-video understanding. VideoZoomer~\cite{VideoZoomer} employs zoom-in attention strategies to selectively process crucial video segments, reducing redundancy while preserving semantic integrity.
Despite these advances in temporal modeling and feature aggregation, most existing approaches still treat reasoning primarily as a textual generation process, without explicitly incorporating visual evidence during intermediate reasoning steps.

\subsection{Multi-modal Chain-of-Thought Reasoning}

With the development of multi-modal large language models~\cite{xie2025ARRA,zheng2025fedvlmbenchbench}, Chain-of-Thought (CoT) prompting~\cite{Chain-of-Thought} introduces explicit intermediate reasoning steps to support multi-step inference. Least-to-most prompting~\cite{zhou2023leasttomostpromptingenablescomplex} further organizes reasoning into ordered sub-problems for complex reasoning tasks. These ideas have been extended to visual reasoning by generating textual reasoning steps conditioned on image or video features. Zebra-CoT~\cite{zebracot} and ThinkMorph~\cite{gu2025thinkmorph} construct large-scale visual-textual interleaved datasets, although their focus is mainly on image reasoning. 

Many works further apply CoT reasoning to video understanding. Video-R1~\cite{Video-R1} combines Chain-of-Thought reasoning with reinforcement learning to improve video reasoning performance. Chain-of-Frames~\cite{Chain-of-Frames} decomposes questions into frame-level reasoning steps, while VideoLLaMA2~\cite{VideoLLaMA2} applies CoT prompting to perform temporal reasoning over encoded video representations. VISTA~\cite{VISTA} implements dynamic CoT routing with self-verification to achieve multi-stage logical decomposition. ViTCoT~\cite{ViTCoT} interleaves video frames and textual reasoning steps, enabling models to repeatedly reference visual content during inference, but it relies on manually selected CoT data and lacks a designed supervision strategy. 

Overall, existing approaches still rely heavily on textual reasoning chains, which limits the effective utilization of video information.
 

\subsection{Contextual Optical Compression(OCR) Rendering}

Recent works~\cite{wu2024deepseekvl2,li2023blip2,alayrac2022flamingo} have explored rendering-based supervision to improve reasoning efficiency by converting structured reasoning into visual representations, which is due to the fact that visual signals possess a significantly higher information density than text. DeepSeek-OCR\cite{DeepSeek-OCR} pioneered the use of CLIP-based feature extraction for document analysis, effectively achieving optical information compression that significantly reduces computational overhead. Render-of-Thought~\cite{Render-of-Thought} introduces rendering of intermediate reasoning steps as images, allowing models to learn from visualized logical trajectories. AgentOCR~\cite{AgentOCR} combines agent-based reasoning with OCR-rendered supervision to improve data compression efficiency. ReGuLaR~\cite{ReGuLaR} employs variational latent reasoning guided by rendered Chain-of-Thought, trying to improve stability in reasoning.

Although these methods demonstrate the potential of rendering-based supervision for efficient reasoning, they are primarily designed for document or text-centric scenarios and have not been fully explored for video reasoning tasks.


\section{Conclusion}

In this work, we present \textbf{VTI-CoT}, a framework for video reasoning that explicitly integrates Chain-of-Thought (CoT) steps with their corresponding visual evidence. We construct visual-textual interleaved CoT instances, rendering interleaved image-text reasoning chains into a canvas. The canvas is then encoded into fixed-length visual features, which serve as the supervision signals for model training. By conducting token-level OCR rendering of visual-textual interleaved content, our method grounds each reasoning step in visual evidence, improving model performance and training efficiency. Extensive experiments on multiple benchmarks, including several long-video datasets, indicating that VTI-CoT outperforms state-of-the-art methods and has faster convergence speed. This framework provides a general paradigm for multi-modal reasoning applicable to instructional video understanding, procedural tasks, and complex multi-agent scenarios.

\clearpage
\bibliographystyle{splncs04}
\bibliography{main}

\clearpage
\appendix

\section{Appendix}






\subsection{Dataset Details}
\label{appendix:dataset}

\begin{figure}[htbp]
    \centering
    \includegraphics[width=0.99\textwidth]{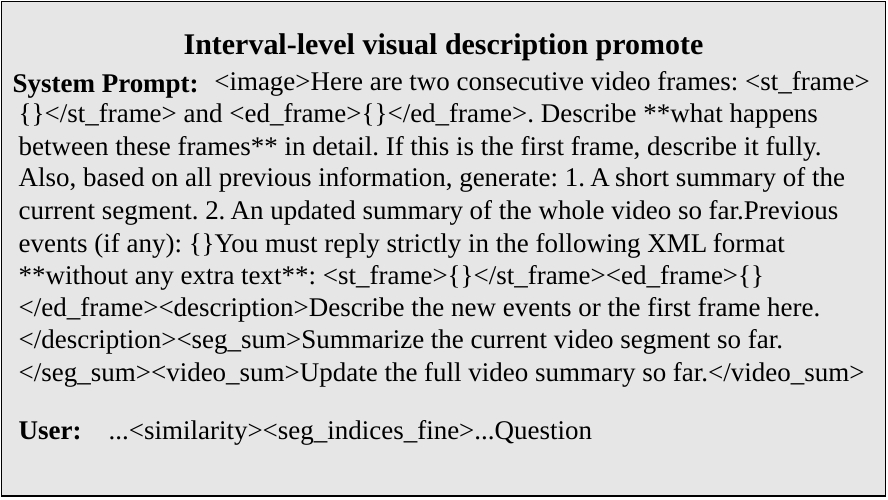}
    \caption{Prompt for interval-level visual description generation.}
    \label{fig:interval_prompt}
\end{figure}

\begin{figure}[htbp]
    \centering
    \includegraphics[width=0.99\textwidth]{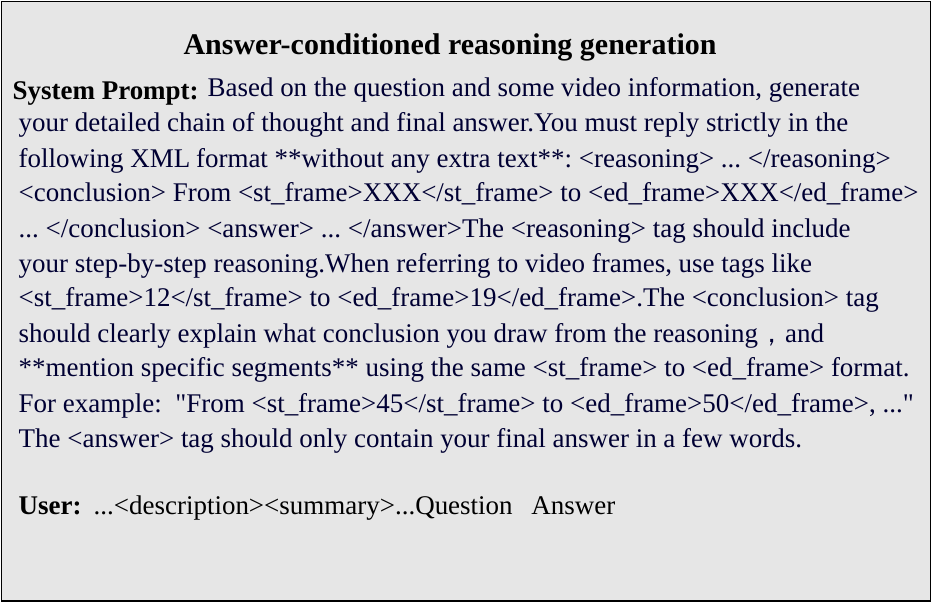}
    \caption{Prompt for answer-conditioned reasoning generation.}
    \label{fig:reasoning_prompt}
\end{figure}

We provide the prompts used for interval-level visual description generation and answer-conditioned reasoning generation in Fig.~\ref{fig:interval_prompt} and Fig.~\ref{fig:reasoning_prompt}. 

\begin{figure}[htbp]
    \centering
    \includegraphics[width=0.99\textwidth]{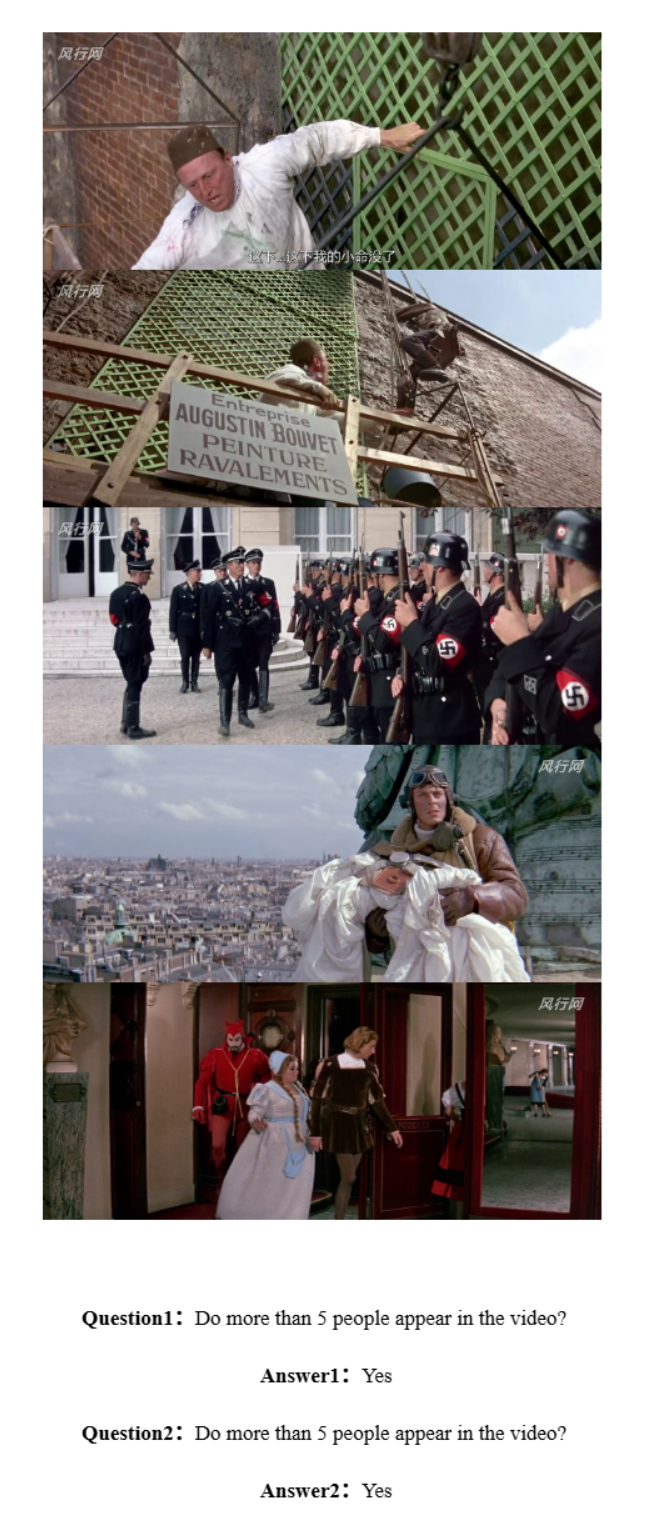}
    \caption{Reasoning example generated by Qwen,Video-R1 and VTI-CoT.}
    \label{fig:reasoning_example}
\end{figure}

\begin{figure}[htbp]
    \centering
    \includegraphics[width=0.55\textwidth]{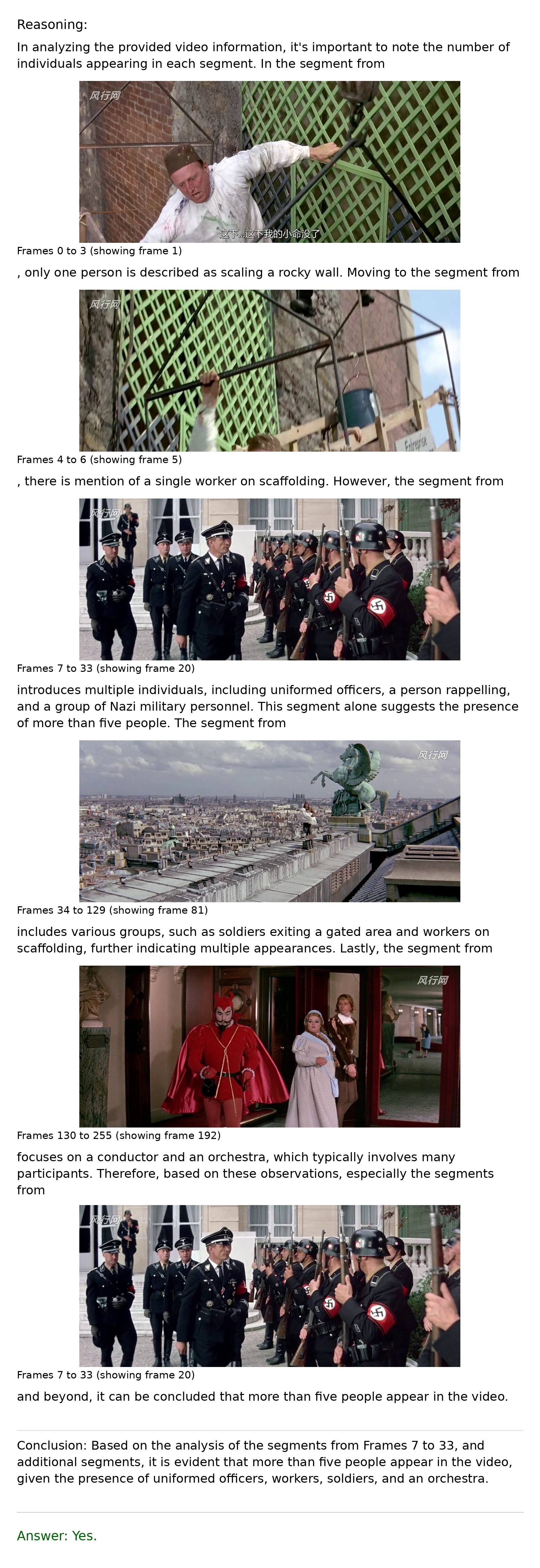}
    \caption{Rendered interleaved CoT canvas.}
    \label{fig:cot_canvas_only}
\end{figure}

Fig.~\ref{fig:reasoning_example} presents an example reasoning instance generated by our VTI-CoT framework on MovieChat using Qwen2.5-VL-7B-Instruct with Video-R1 cold-start data. It shows the sampled video frames along with the associated question and answer.  
Fig.~\ref{fig:cot_canvas_only} illustrates the corresponding visual-text interleaved CoT canvas, which serves as the compact supervision signal for training.

\subsection{Training Details}
\label{appendix:train}

Our training consists of a supervised fine-tuning (SFT) stage. Key hyperparameters are summarized in Table~\ref{tab:training_hyperparameters}. During training, the vision encoder is frozen while optimizing the language backbone. The learning rate follows a linear decay schedule. Training is conducted on multiple GPUs and typically finishes within several hours depending on the dataset and stage.

\paragraph{Network Architecture.}  
We adopt the Qwen2.5-VL-7B model as the backbone of our framework. This model consists of two main components:
\begin{itemize}
    \item LLM Backbone: A 7B-parameter autoregressive transformer with 28 layers, hidden size 3584, and 28 attention heads. It is responsible for generating multi-step CoT reasoning and final answers.
    \item Vision Encoder: A transformer-based module that extracts patch-level features from video frames (28×28 patches) and projects them into the language hidden space. Cross-attention layers in the language model allow fusion of visual and textual information.
\end{itemize}

\paragraph{Design Rationale.}  
We choose Qwen2.5-VL-7B due to its strong visual-textual alignment capabilities and capacity to model complex reasoning chains. The large language backbone allows the model to handle long CoT sequences, while the patch-level vision encoder preserves spatial details essential for grounded video reasoning. Freezing the vision encoder stabilizes visual features, reduces training complexity, and focuses optimization on the language backbone. This design ensures that the model can accurately capture and reason over visual cues, improving CoT reasoning and inference accuracy.

\paragraph{Training Protocol.}  
We employ a batch size of 1, training for 1 epoch. The learning rate is set to $3\times10^{-6}$ with linear decay. AdamW optimizer with weight decay is used to stabilize training. Multi-GPU training is leveraged to accelerate computation while maintaining batch size constraints due to memory limits from video input and the large language model.

\begin{table}[htbp]
\centering
\caption{Key training hyperparameters and network configuration used in our experiments.}
\label{tab:training_hyperparameters}
\begin{tabular}{l c}
\hline
\textbf{Hyperparameter / Component} & \textbf{Value} \\
\hline
Train epochs & 1 \\
Train batch size & 1 \\
Learning rate & $3\times10^{-6}$ \\
Learning rate scheduler & Linear \\
Optimizer & AdamW \\ 
Freeze vision encoder & True \\
Backbone & Qwen2.5-VL-7B \\
Number of layers & 28 \\
Hidden size & 3584 \\
Number of attention heads & 28 \\
Patch size & 28$\times$28 \\
\hline
\end{tabular}
\end{table}






\subsection{Additional Experiments}

\paragraph{Effect of Visual Feature Alignment Layers} We study how the injection depth of visual features affects reasoning performance. Visual tokens are aligned to four layers of the Transformer: first, middle, penultimate, and final. As shown in Table~\ref{tab:alignment_layer}, aligning features at the final layer consistently achieves the best results across all benchmarks. The penultimate layer performs slightly worse, while middle and first layers show larger drops, especially on long-video reasoning tasks. This suggests that fusing visual information too early can limit the model's ability to form stable, high-level reasoning representations. High-level semantic features at the final layer provide richer temporal and causal abstractions, supporting more accurate multi-step reasoning.

\begin{table}[htbp]
\centering
\caption{Experiment on visual feature alignment at different Transformer layers.
Visual tokens are injected at the first layer, a middle layer, the penultimate layer, and the final layer. Final-layer alignment consistently yields the best results.}
\label{tab:alignment_layer}
\resizebox{\linewidth}{!}{
\begin{tabular}{l|c|c|c|c|c|c}
\toprule
\textbf{Alignment Layer} 
& MVBench 
& TempCompass 
& Video-MME 
& MMVU 
& LVBench 
& LongVideoBench \\
\midrule
Final layer
& \textbf{65.9} & \textbf{74.5} & \textbf{59.6} & \textbf{65.3} & \textbf{40.5} & \textbf{55.0} \\
\midrule
Penultimate layer 
& 65.5 & 73.5 & 57.7 & 64.9 & 37.8 & 53.0 \\
Middle layer 
& 65.4 & 73.0 & 57.3 & 65.0 & 37.2 & 52.3 \\
First layer  
& 65.3 & 73.3 & 57.4 & 64.6 & 36.0 & 52.5 \\
\bottomrule
\end{tabular}
}
\end{table}

\end{document}